\documentclass[pdflatex,sn-mathphys-num]{sn-jnl}

\usepackage{graphicx}%
\usepackage{multirow}%
\usepackage{amsmath,amssymb,amsfonts}%
\usepackage{amsthm}%
\usepackage{mathrsfs}%
\usepackage[title]{appendix}%
\usepackage{xcolor}%
\usepackage{textcomp}%
\usepackage{manyfoot}%
\usepackage{booktabs}%
\usepackage{algorithm}%
\usepackage{algorithmicx}%
\usepackage{algpseudocode}%
\usepackage{listings}%

\usepackage{makecell}
\usepackage{silence}
\usepackage[switch]{lineno}
\usepackage[utf8]{inputenc}
\usepackage{array}
\usepackage{siunitx}
\usepackage{ulem}

\begin{document}


\title[Article Title]{SlingBAG Pro: Accelerating point cloud-based iterative
reconstruction for 3D photoacoustic imaging with arbitrary array
geometries}








\author[1]{\fnm{Shuang} \sur{Li}}

\author[1]{\fnm{Yibing} \sur{Wang}}

\author[2]{\fnm{Jian} \sur{Gao}}

\author[3]{\fnm{Chulhong} \sur{Kim}}

\author[3]{\fnm{Seongwook} \sur{Choi}}

\author[1]{\fnm{Yu} \sur{Zhang}}

\author[1]{\fnm{Qian} \sur{Chen}}

\author*[2]{\fnm{Yao} \sur{Yao}}\email{yaoyao@nju.edu.cn}

\author*[1, 4]{\fnm{Changhui} \sur{Li}}\email{chli@pku.edu.cn}

\affil*[1]{\orgdiv{Department of Biomedical Engineering, College of Future Technology}, \orgname{Peking University}, \city{Beijing}, \country{China}}

\affil*[2]{\orgdiv{School of Intelligence Science and Technology}, \orgname{Nanjing University}, \city{Suzhou}, \country{China}}

\affil[3]{\orgdiv{Department of Electrical Engineering, Convergence IT Engineering, Mechanical Engineering, and Medical Science and Engineering, Medical Device Innovation Center}, \orgname{Pohang University of Science and Technology}, \city{Pohang}, \country{Republic of Korea}}

\affil*[4]{\orgdiv{National Biomedical Imaging Center}, \orgname{Peking University}, \city{Beijing}, \country{China}}

\abstract{High-quality three-dimensional (3D) photoacoustic imaging (PAI) is gaining increasing attention in clinical applications. To address the challenges of limited space and high costs, irregular geometric transducer arrays that conform to specific imaging regions are promising for achieving high-quality 3D PAI with fewer transducers. However, traditional iterative reconstruction algorithms struggle with irregular array configurations, suffering from high computational complexity, substantial memory requirements, and lengthy reconstruction times. In this work, we introduce SlingBAG Pro, an advanced reconstruction algorithm based on the point cloud iteration concept of the Sliding ball adaptive growth (SlingBAG) method, while extending its compatibility to arbitrary array geometries. SlingBAG Pro maintains high reconstruction quality, reduces the number of required transducers, and employs a hierarchical optimization strategy that combines zero-gradient filtering with progressively increased temporal sampling rates during iteration. This strategy rapidly removes redundant spatial point clouds, accelerates convergence, and significantly shortens overall reconstruction time. Compared to the original SlingBAG algorithm, SlingBAG Pro achieves up to a 2.2-fold speed improvement in point cloud-based 3D PA reconstruction under irregular array geometries. The proposed method is validated through both simulation and in vivo mouse experiments, and the source code is publicly available at \href{https://github.com/JaegerCQ/SlingBAG_Pro}{https://github.com/JaegerCQ/SlingBAG\_Pro}.}


\maketitle

\section{Introduction}\label{sec1}

Photoacoustic imaging (PAI) is a hybrid imaging modality that leverages ultrasound detection alongside optical absorption contrast, enabling non-invasive visualization of biological tissues at centimeter-scale depths with spatial resolutions superior to conventional optical imaging techniques. Due to these advantages, PAI has found wide application in both preclinical studies and clinical settings~\cite{park2024clinical, wang2012photoacoustic, assi2023review, dean2017advanced, lin2022emerging, ntziachristos2024addressing}. Advancements in three-dimensional (3D) PAI using 2D matrix arrays have enabled in vivo imaging of peripheral vessels~\cite{wray2019photoacoustic, li2024photoacoustic}, breast tissues~\cite{wang2021functional, han2021three}, and small animals~\cite{sun2024real} with promising results. Various array designs, including spherical, planar~\cite{matsumoto2018label, matsumoto2018visualising, ivankovic2019real, dean2013portable, nagae2018real, kim2023wide, piras2009photoacoustic, heijblom2012visualizing}, and synthetically scanned arrays~\cite{li2024photoacoustic, wang2024comprehensive}, have been explored to enhance 3D imaging capabilities.

However, conventional hemispherical and planar arrays are often bulky and limited by their angular coverage, necessitating rigid subject positioning during imaging and restricting time-resolved 3D PAI especially in free-moving or patient-conformal scenarios. Solutions such as miniaturized, irregular arrays and integration with piezoelectric micromachined ultrasonic transducers (PMUTs), capacitive micromachined ultrasonic transducers (CMUTs), or flexible materials have been investigated to provide greater conformity to target anatomies and to enable dynamic 3D imaging. Yet, arbitrary and sparsely arranged arrays with irregular geometries and complex transducer orientations are significantly challenging for traditional universal back-projection (UBP) algorithms, making high-quality reconstruction reliant on advanced iterative algorithms.

Iterative reconstruction (IR) methods have shown promise in handling arbitrary and sparse array configurations for PAI. Early approaches, such as that by Paltauf et al.~\cite{paltauf2002iterative}, minimized discrepancies between measured and simulated signals iteratively, while later works incorporated more sophisticated physical modeling~\cite{dean2012accurate, wang2012investigation, wang2014discrete, huang2013full, zhu2023mitigating, huynh2024fast, treeby2010k}. Although these IR methods offer excellent image quality, their computational and memory demands grow rapidly with the grid size and the complexity of 3D datasets—rendering large-scale applications impractical in many cases.

To address these limitations, model-based IR techniques using semi-analytical forward models and on-the-fly computation~\cite{ding2017efficient, ding2020model, rosenthal2010fast, pandey2023model, paige1982lsqr} have been developed to bypass explicit system matrix storage. Nonetheless, these methods often rely on approximations that can degrade accuracy for near-field or complex geometries, and their dependence on solvers like LSQR makes it difficult to incorporate advanced non-smooth regularization (e.g., $L_1$ or TV norms). Additionally, their uniform voxel-grid representation is inherently inefficient for reconstructing sparse and structurally complex features, such as vascular networks, due to the trade-off between resolution and computational cost.

Recently, the sliding Gaussian ball adaptive growth (SlingBAG) algorithm~\cite{li2024sliding} introduced a point cloud-based iterative reconstruction paradigm, which notably alleviates memory constraints and enables high-quality 3D PAI reconstruction for large-scale datasets. However, the original SlingBAG framework is limited to regular imaging regions and requires a large, randomly initialized point cloud, resulting in redundancies and suboptimal computational efficiency especially when applied to miniaturized or wearable systems with irregularly arranged transducers.

To overcome these challenges, we propose SlingBAG Pro, an advanced iterative reconstruction framework that generalizes the point cloud approach to arbitrary geometric array configurations. SlingBAG Pro incorporates a zero-gradient filtering process to optimize point cloud initialization, reducing the number of points while improving their correspondence with true photoacoustic source distributions. Moreover, a hierarchical optimization strategy involving progressively increasing temporal sampling rates significantly accelerates point cloud convergence during iteration and removes redundant points efficiently.

In this work, we systematically present the SlingBAG Pro algorithm, detailing its flexibility for diverse array geometries, zero-gradient filtering, and hierarchical optimization techniques. Through comprehensive simulations and in vivo experimental validation, we demonstrate that SlingBAG Pro delivers high-quality, efficient 3D PAI, paving the way for its application in clinical and wearable photoacoustic imaging scenarios with arbitrary sparse arrays.

\section{Results}\label{sec2}

\subsection{3D PA image reconstruction under sparse irregular array}\label{subsec2_1}

\begin{figure}[htbp]
\centering
\includegraphics[width=1.0\textwidth]{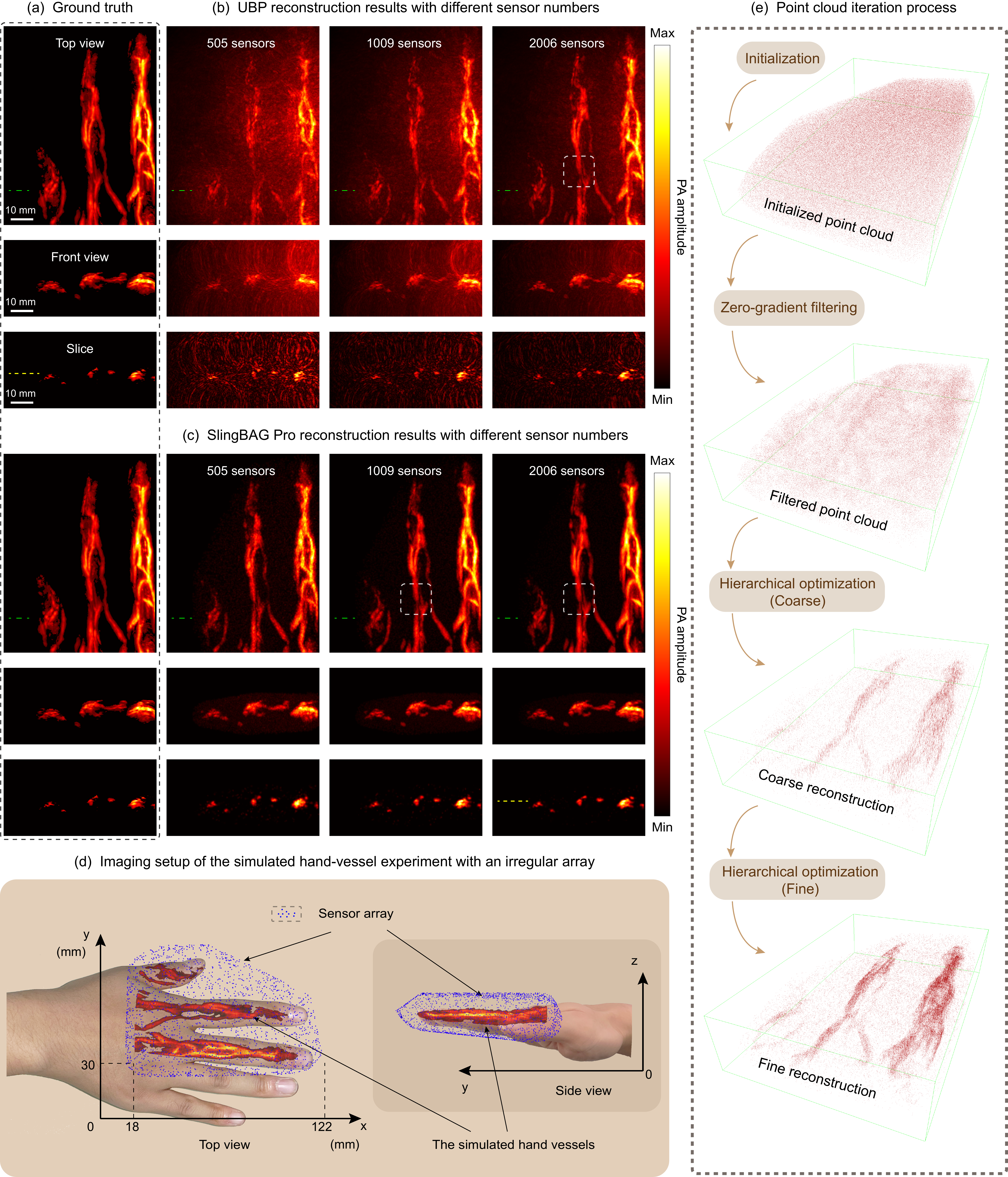}
\caption{Comparison of 3D photoacoustic reconstruction results under sparse irregular arrays. (a) Top-view maximum amplitude projection (MAP), front-view MAP, and the cross-sectional slice along the green dashed line in the top-view MAP of the acoustic source. (b) Top-view MAP, front-view MAP, and corresponding cross-sectional slice along the green dashed line in the top-view MAP of the UBP reconstruction results with 505, 1009, 2006 sensors, respectively. (c) Top-view MAP, front-view MAP, and corresponding cross-sectional slice along the green dashed line in the top-view MAP of the SlingBAG Pro reconstruction results with 505, 1009, 2006 sensors, respectively. (d) Imaging setup. (e) Point cloud iteration process of SlingBAG Pro reconstruction with 2006 sensors. (Scale bar: 10 mm.)}\label{fig1}
\end{figure}

We first designed simulation experiments to verify the superior reconstruction capability and speed improvement of the SlingBAG Pro algorithm under arbitrary irregular array configurations. The simulation uses 3D hand vessels as the reconstruction target. The 3D hand vessels simulation data is based on the 3D PA imaging results of human peripheral hand vessels from previous work~\cite{li2024photoacoustic}, and is used as the ground truth (Fig. \ref{fig1}(a)). With a grid point spacing of 0.2 mm, the total grid size of the imaging area is $400 \times 520 \times 200$ ($80\ \text{mm} \times 104\ \text{mm} \times 40\ \text{mm}$). The simulated hand vessels are located inside the imaging region, and the irregular detection array is arranged around the vessels (Fig. \ref{fig1}(d)), forming a closed envelope. We set different numbers of detectors on the envelope to study the 3D PA reconstruction under sparse and irregular array conditions, with the number of detectors set to 505, 1009, and 2006, randomly and uniformly sampled on the envelope. In the simulation experiments, the sampling frequency is 40 MHz, with 4900 sampling points, a medium density of \(1 \times 10^3\) kg/m³, and a speed of sound of 1500 m/s.

\begin{figure}[htbp]
\centering
\includegraphics[width=1.0\textwidth]{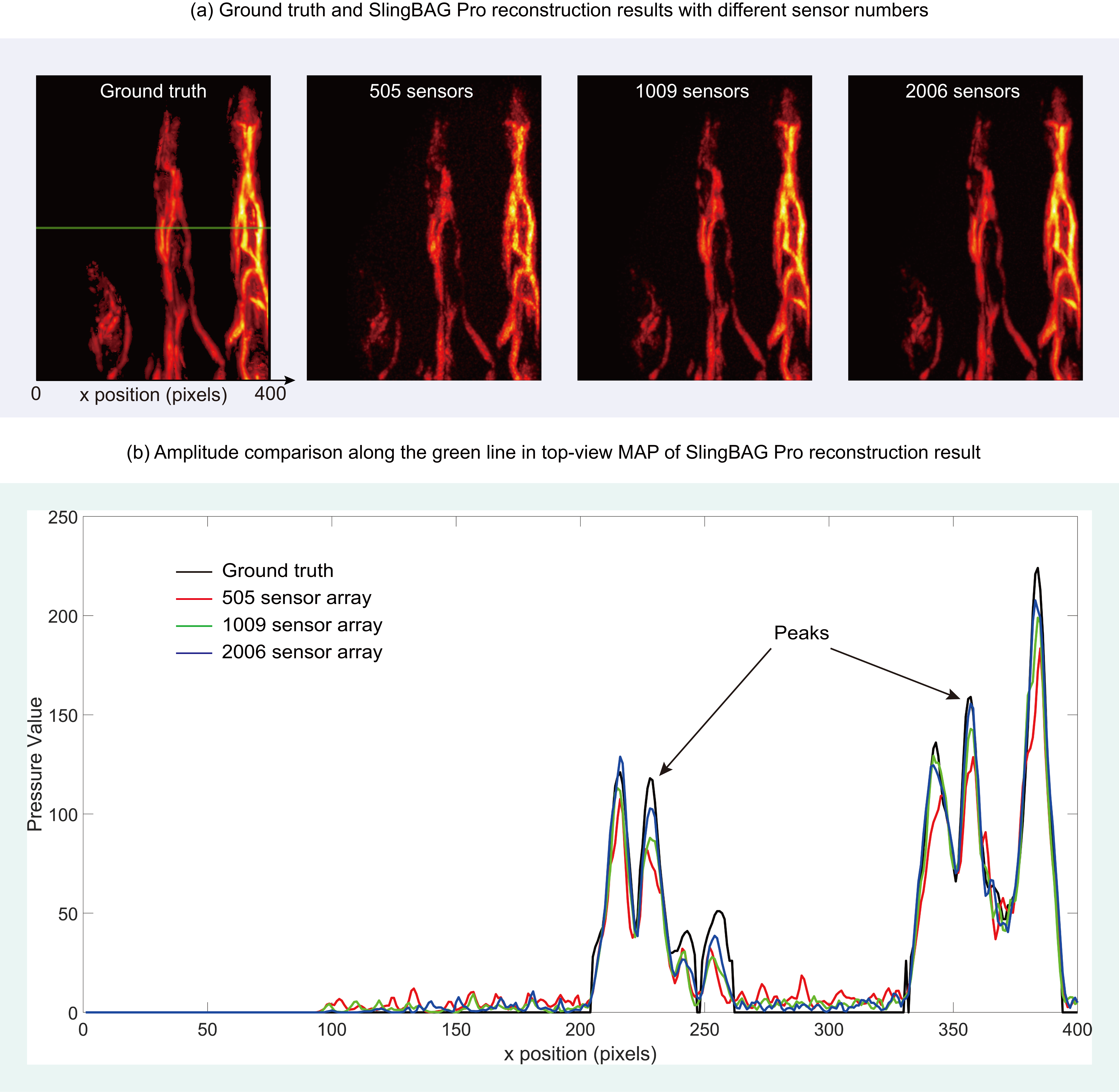}
\caption{Quantitative accuracy assessment of SlingBAG Pro reconstruction results. (a) Top-view maximum amplitude of Ground truth and SlingBAG Pro reconstruction results with different sensor numbers. (b) Amplitude comparison along the green line in top-view MAP of SlingBAG Pro reconstruction result.}\label{fig2}
\end{figure}

Based on the detector coordinates, SlingBAG Pro constructs an envelope surface corresponding to the irregular array and then defines the reconstruction region inward along the normal direction of the envelope. For the three experimental conditions with 505, 1009, and 2006 detectors, SlingBAG Pro used 200,000 points to initialize the reconstruction space point cloud. The reconstruction results of all three experiments are shown in Fig. \ref{fig1}, with UBP reconstruction results included for comparison. It is evident that SlingBAG Pro demonstrates unparalleled 3D reconstruction capability under sparse and irregular array configurations. For extremely sparse arrays with 505 and 1009 sensors, the UBP results are dominated by severe reconstruction artifacts, making it nearly impossible to distinguish the vascular structures (Fig. \ref{fig1}(b)). In contrast, even with just 505 sensors, SlingBAG Pro achieves very high reconstruction quality; as the number of detectors increases, the reconstruction fidelity and detail recovery of SlingBAG Pro further improve. When the number of sensors increases from 1009 to 2006, the vessel details within the white dashed box become even clearer (Fig. \ref{fig1}(c)). Furthermore, we quantitatively evaluated the reconstruction quality of SlingBAG Pro using Mean Squared Error (MSE), Peak Signal-to-Noise Ratio (PSNR), and Structural Similarity Index Measure (SSIM) as metrics (Tab. \ref{tab1}). For the case with 505 sensors, the PSNR of SlingBAG Pro reconstruction reached 26.44 dB, which is much higher than the UBP result with 2006 sensors (which only reached a PSNR of 21.95 dB). As the number of sensors increases, the PSNR of SlingBAG Pro's reconstruction further improves, reaching 29.22 dB for 2006 detectors, and SSIM increases from 0.565 for 505 sensors to 0.722 for 2006 sensors. Furthermore, we compared the unnormalized pixel values along the green line in the top-view maximum amplitude projection of the reconstruction result with those of the ground truth (Fig. \ref{fig2}(a)) to investigate the quantitative accuracy of SlingBAG Pro reconstruction. As shown in Fig. \ref{fig2}(b), the reconstructed pixel values of SlingBAG Pro agree well overall with those of the ground truth. Moreover, as the number of detectors increases from 505 to 1009 and then to 2006, the agreement between the "peaks" in the reconstruction and those in the ground truth improves further, demonstrating the quantitative accuracy of the SlingBAG Pro algorithm.

\renewcommand{\arraystretch}{1.4}
\begin{table}[htbp]
    \centering
    \caption{Quantitative evaluation for reconstruction results of hand vessels from different algorithms}
    \begin{tabular}{c|c|c|c|c|c}
      \hline
      \textbf{Sensor Number} & \textbf{Algorithm} & \textbf{Time (h)} & \textbf{MSE $\downarrow$} & \textbf{PSNR (dB) $\uparrow$} & \textbf{SSIM $\uparrow$} \\
      \hline
      \multirow{3}{*}{2006}
      & SlingBAG Pro & \textbf{3.79}  & \textbf{0.00120} & \textbf{29.22} & \textbf{0.7223} \\
      & SlingBAG      & 8.32  & 0.00130 & 28.86 & 0.6519 \\
      & UBP           & --    & 0.00638 & 21.95 & 0.2157 \\
      \hline
      \multirow{3}{*}{1009}
      & SlingBAG Pro & \textbf{2.47}  & \textbf{0.00152} & \textbf{28.20} & \textbf{0.6379} \\
      & SlingBAG      & 4.23  & 0.00161          & 27.92          & 0.5892 \\
      & UBP           & --    & 0.00884          & 20.54          & 0.1808 \\
      \hline
      \multirow{3}{*}{505}
      & SlingBAG Pro & \textbf{1.71}  & \textbf{0.00227} & \textbf{26.44} & \textbf{0.5654} \\
      & SlingBAG      & 2.25  & 0.00230          & 26.38          & 0.5300 \\
      & UBP           & --    & 0.01334          & 18.75          & 0.1436 \\
      \hline
    \end{tabular}
    \label{tab1}
\end{table}

We then compared the reconstruction speed and quality of SlingBAG Pro and SlingBAG, as shown in Tab. \ref{tab1}. It is evident that, due to the adoption of zero-gradient filtering and the hierarchical optimization strategy, SlingBAG Pro offers substantial improvements in reconstruction speed over the original SlingBAG algorithm—an advantage that becomes even more pronounced for large-scale reconstructions. For example, with 2006 detectors, when the PSNR reaches a comparable level (29.22 dB for SlingBAG Pro versus 28.86 dB for SlingBAG), SlingBAG Pro requires only 3.79 hours, representing a 2.2-fold speed improvement over the original SlingBAG algorithm (which takes 8.32 hours). Similar significant speed improvements are observed for reconstructions with 505 and 1009 detectors. Fig. \ref{fig1}(e) illustrates the point cloud iteration process of SlingBAG Pro reconstruction with 2006 detectors. As shown, after randomly and uniformly initializing the point cloud in the reconstruction region (with 200,000 points), application of zero-gradient filtering reduces the number of points to 106,924, revealing a very rough outline of the hand vasculature. Next, through coarse reconstruction based on the hierarchical optimization strategy, SlingBAG Pro yields a sparser point cloud representation of the hand vessels. In the final stage, fine reconstruction, which is also based on hierarchical optimization, activates point duplication and iterative position optimization, ultimately yielding a fine-grained point cloud reconstruction result with accurate structural information. For the hand simulation experiments, SlingBAG Pro adopted a hierarchical optimization scheme with successive downsampling, progressing from 16× downsampling to 4× downsampling and finally to the original sampling rate.

\begin{figure}[htbp]
\centering
\includegraphics[width=1.0\textwidth]{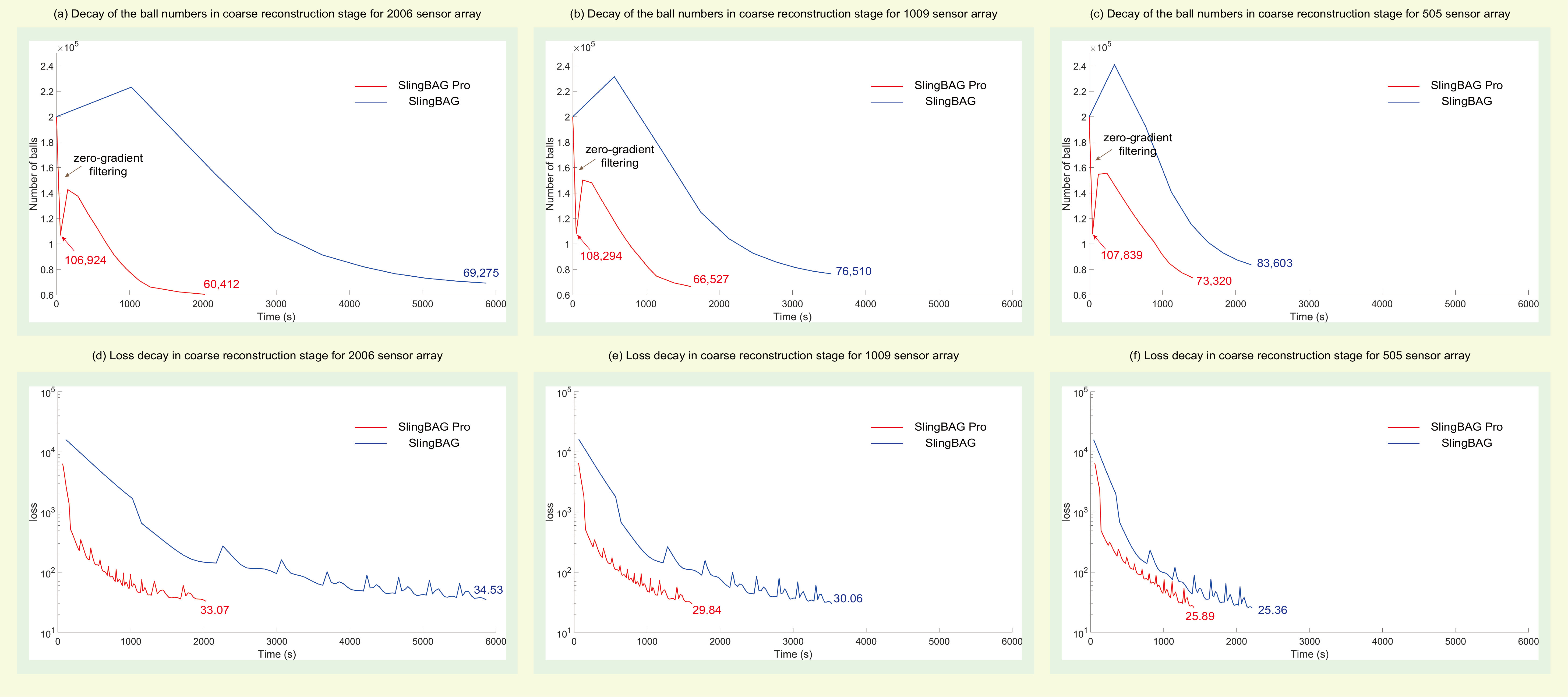}
\caption{Comparison of decay for both the ball numbers and the loss in coarse reconstruction stage between SlingBAG Pro and SlingBAG. (a-c) Decay of the ball numbers in coarse reconstruction stage for 505, 1009, 2006 sensor array, respectively. (d-f) Loss decay in coarse reconstruction stage for 505, 1009, 2006 sensor array, respectively.}\label{fig3}
\end{figure}

Additionally, we separately compared the convergence of the number of balls and the loss over time during the coarse reconstruction stage for both SlingBAG Pro and the original SlingBAG algorithm (Fig. \ref{fig3}). With the same initialization of 200,000 balls, zero-gradient filtering rapidly reduces the number of balls in the point cloud (as indicated by the brown arrows in Fig. \ref{fig3}(a-c)), resulting in a much lower initial loss for SlingBAG Pro compared to the original SlingBAG with random initialization, thereby improving reconstruction efficiency. Throughout the entire coarse reconstruction process, the learning rates for all parameters in SlingBAG Pro and SlingBAG were kept the same. It is evident that the adoption of hierarchical optimization in SlingBAG Pro leads to much faster convergence in both the number of balls and the loss, especially in the case of 2006 detectors, where SlingBAG Pro achieves in 2,000 seconds what SlingBAG requires 6,000 seconds to accomplish. This demonstrates the remarkable speed advantage of SlingBAG Pro.

\begin{figure}[htbp]
\centering
\includegraphics[width=1.0\textwidth]{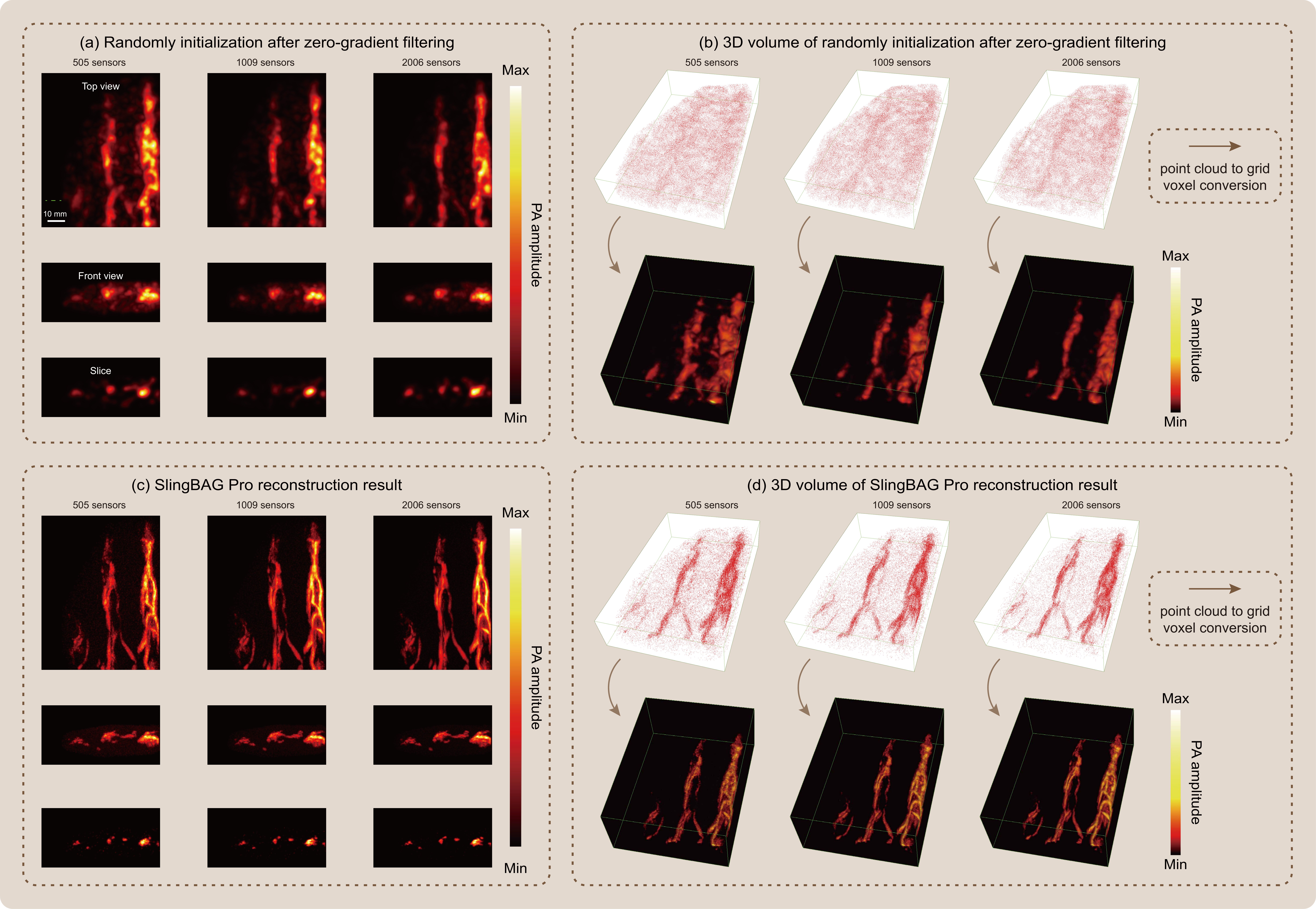}
\caption{Comparison of 3D photoacoustic reconstruction results under sparse irregular arrays. (a) Top-view maximum amplitude projection (MAP), front-view MAP, and the cross-sectional slice along the green dashed line in the top-view MAP for 3D volume of randomly initialization after zero-gradient filtering with 505, 1009, 2006 sensors, respectively. (b) Point cloud result and 3D volume of randomly initialization after zero-gradient filtering with 505, 1009, 2006 sensors, respectively. (c) Top-view MAP, front-view MAP, and corresponding cross-sectional slice along the green dashed line in the top-view MAP of the SlingBAG Pro reconstruction results with 505, 1009, 2006 sensors, respectively. (d) Point cloud result and 3D volume of SlingBAG Pro reconstruction with 505, 1009, 2006 sensors, respectively. (Scale bar: 10 mm.)}\label{fig4}
\end{figure}

Furthermore, we demonstrate the distribution of point clouds after zero-gradient filtering, as well as the numerical distribution of zero-valued gradients within the reconstruction region using the SlingBAG Pro algorithm (Fig. \ref{fig4}). For reconstructions with 505, 1009, and 2006 detectors, zero-gradient computation was performed at 16× downsampling—the same as the lowest sampling rate in the hierarchical optimization. After temporarily setting the initial pressure of all 200,000 initialized points to zero, we performed a single gradient computation and retained all points with negative initial pressure gradients. The absolute value of the initial pressure gradient was then assigned as the intensity of these retained points, which were subsequently rendered using the point cloud-to-voxel shader (Fig. \ref{fig4}(b)). By comparing the post-filtered point cloud and voxel results with those after full iterative optimization (Fig. \ref{fig4}(d)), it can be seen that the retained points accurately outline the hand vessel structures. As the number of detectors increases from 505 to 1009 to 2006, the point clouds retained after zero-gradient filtering more clearly delineate the hand vessel contours. Fig. \ref{fig4}(a)(c) show the maximum intensity projections (MAP) of the 3D volumes: the gradient-filtered pressure gradient volume and the final reconstructed volume, respectively. Under the 505-detector condition, the filtered pressure gradient volume displays a blurrier and noisier outline, whereas with 2006 detectors, the noise is substantially reduced, resulting in a high-quality initial vessel contour.

\subsection{3D PA image reconstruction of in vivo animal studies with hierarchical optimization}\label{subsec2_2}

Subsequently, we reconstructed images of real rat experimental data utilizing both the UBP and SlingBAG Pro algorithms, and compared the results with those from the original SlingBAG algorithm to demonstrate the computational efficiency advantage of SlingBAG Pro. The in vivo animal data comprising rat kidney and rat liver were collected by Kim’s lab using a hemispherical ultrasound (US) transducer array with 1024 elements and a 60 mm radius~\cite{choi2023deep}. Each element in the array had an average center frequency of 2.02 MHz and a bandwidth of 54\%. The system provided an effective field of view (FOV) of $12.8 \text{ mm} \times 12.8 \text{ mm} \times 12.8 \text{ mm}$, and the spatial resolution was nearly isotropic in all directions at approximately 380 µm when all 1024 elements were active~\cite{choi2023deep}. All in vivo measurements were performed with a sampling frequency of 8.33 MHz and included 896 sampling points. Further details regarding the 3D PAI system and animal experiments can be found in the literature~\cite{choi2023deep, choi2023recent, kim20243d, kim2022deep, Kim2024inpress}.

\begin{figure}[htbp]
\centering
\includegraphics[width=1.0\textwidth]{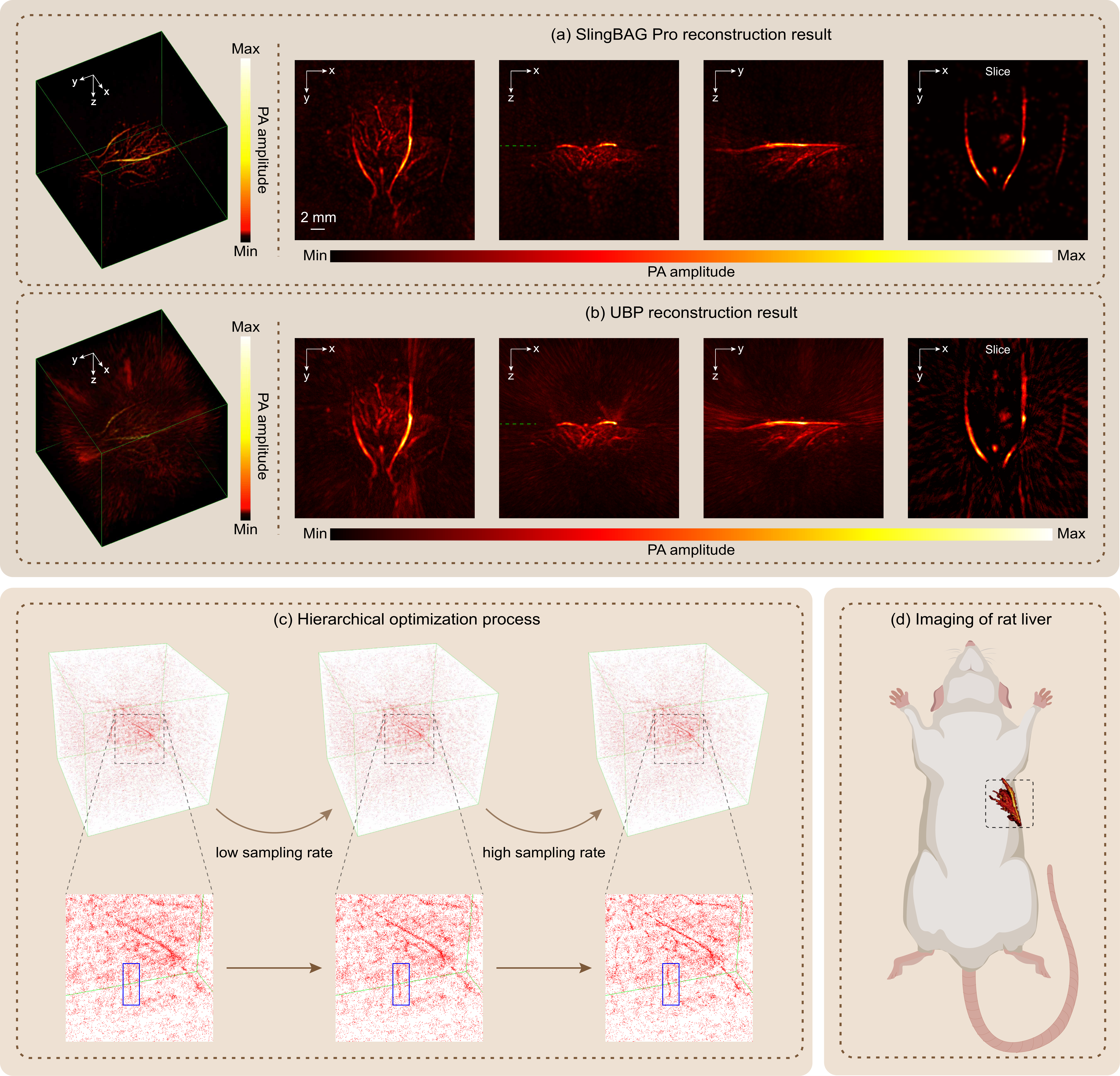}
\caption{3D PA reconstruction results of a rat liver. (a) 3D volume, XY Plane-MAP, XZ Plane-MAP, YZ Plane-MAP and the cross-section slice at green dashed line marked in XZ Plane-MAP of the SlingBAG Pro 3D reconstruction results using 1024 sensor signals. (b) XY Plane-MAP, XZ Plane-MAP, YZ Plane-MAP and the cross-section slice at green dashed line marked in XZ Plane-MAP of the UBP 3D reconstruction results using 1024 sensor signals. (Scale: 2 mm.) (c) Hierarchical optimization process of point cloud iterative reconstruction. (d) Schematic diagram of the imaging area.}\label{fig5}
\end{figure}

\begin{figure}[htbp]
\centering
\includegraphics[width=1.0\textwidth]{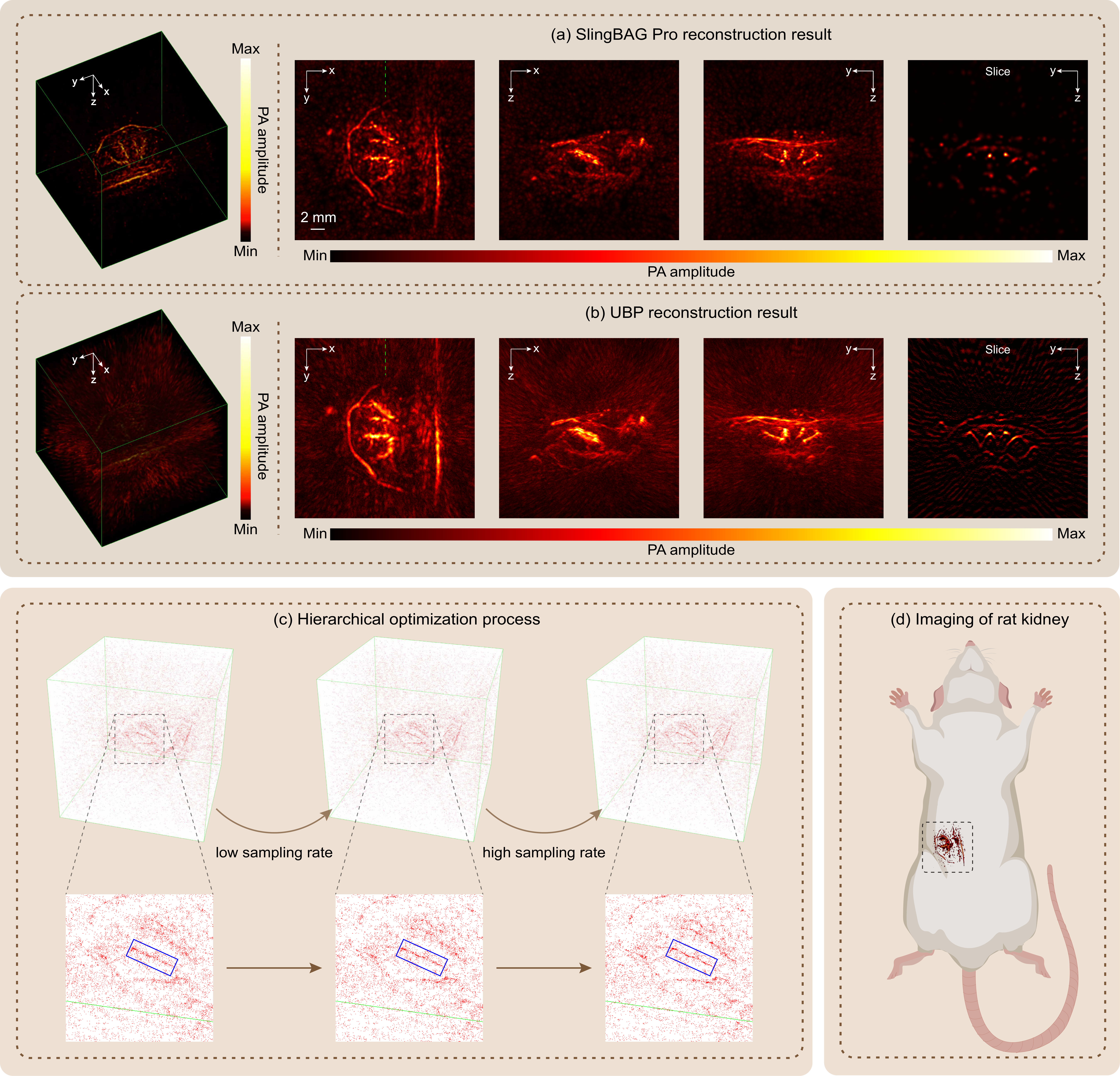}
\caption{3D PA reconstruction results of a rat kidney. (a) 3D volume, XY Plane-MAP, XZ Plane-MAP, YZ Plane-MAP and the cross-section slice at green dashed line marked in XY Plane-MAP of the SlingBAG Pro 3D reconstruction results using 1024 sensor signals. (b) XY Plane-MAP, XZ Plane-MAP, YZ Plane-MAP and the cross-section slice at green dashed line marked in XY Plane-MAP of the UBP 3D reconstruction results using 1024 sensor signals. (Scale: 2 mm.) (c) Hierarchical optimization process of point cloud iterative reconstruction. (d) Schematic diagram of the imaging area.}\label{fig6}
\end{figure}

\begin{table}[htbp]
    \centering
    \caption{CNR and reconstruction time for liver and kidney with different methods}
    \begin{tabular}{c|c|c|c|c}
        \hline
        \textbf{In Vivo Experiments} & \textbf{Method} & \textbf{CNR $\uparrow$} & \textbf{Time (h)} & \textbf{Time (s)} \\
        \hline
        \multirow{3}{*}{Rat Liver}
        & Sling BAG Pro & 44.6882 & \textbf{3.37} & 12130 \\
        & SlingBAG      & 42.0117          & 4.75 & 17117 \\
        & UBP           & 30.6188          & -- & -- \\
        \hline
        \multirow{3}{*}{Rat Kidney}
        & Sling BAG Pro & 28.5127          & \textbf{1.81} & 6522 \\
        & SlingBAG      & 30.2625 & 2.90 & 10455 \\
        & UBP           & 20.1798           & -- & -- \\
        \hline
    \end{tabular}
    \label{tab2}
\end{table}

We further present the reconstruction results for rat liver (Fig. \ref{fig5}) and rat kidney (Fig. \ref{fig6}). It can be seen that SlingBAG Pro demonstrates unmatched high-quality reconstruction capability. From the 3D volumes of liver (Fig. \ref{fig5}(a)) and kidney (Fig. \ref{fig6}(a)) reconstructed by SlingBAG Pro, the resulting images contain minimal artifacts. In contrast, the 3D volumes reconstructed by the UBP algorithm for liver (Fig. \ref{fig5}(b)) and kidney (Fig. \ref{fig6}(b)) exhibit extremely severe artifacts; without maximum amplitude projection, the vasculature is almost completely invisible in the UBP-reconstructed 3D volumes. The slice images of the liver and kidney reconstructions further illustrate this difference: UBP results show extremely strong radial and arc-shaped artifacts, while SlingBAG Pro exhibits minimal artifacts and high-quality structural recovery. We also show the intermediate point cloud results of the hierarchical optimization used by SlingBAG Pro for the liver (Fig. \ref{fig5}(c)) and kidney (Fig. \ref{fig6}(c)). For in vivo experiments, SlingBAG Pro employed a hierarchical strategy from 4× downsampling to 2× downsampling to full sampling rate. As highlighted by the blue rectangles, vessel structures in the point cloud become concentrated and clear as the sampling rate increases through the stages of hierarchical optimization. Table \ref{tab2} presents the comparison of CNR and reconstruction time for different methods. It can be seen that SlingBAG Pro reconstructions have much higher CNR than UBP. Moreover, while maintaining the high reconstruction quality of SlingBAG, SlingBAG Pro substantially improves reconstruction speed: for rat kidney reconstruction, SlingBAG required 2.90 hours, whereas SlingBAG Pro completed the reconstruction in only 1.81 hours, reducing the reconstruction time by more than 1 hour.

\section{Discussion}\label{sec3}

In this work, we propose the SlingBAG Pro algorithm for large-scale 3D photoacoustic iterative reconstruction under arbitrary geometric array configurations. Building upon the SlingBAG point cloud iterative framework, SlingBAG Pro optimizes for arbitrary array geometries, substantially improves point cloud initialization through zero-gradient filtering, and greatly enhances point cloud-based iterative reconstruction efficiency via hierarchical optimization strategies. 

In simulation studies, visualization of the point cloud after zero-gradient filtering and its conversion to a voxel grid demonstrates that SlingBAG Pro effectively optimizes the distribution of randomly initialized points using zero-gradient filtering. The filtered results (Fig. \ref{fig4}(a-b)) form a well-defined vessel distribution contour that closely resembles the final reconstruction outcome (Fig. \ref{fig4}(c-d)), supporting the reliability of the filtered point cloud as an initial estimate and providing a solid foundation for further point cloud iteration. This optimized point cloud initialization method is perfectly compatible with the point cloud iterative framework, which provides a universal and efficient initialization strategy for point cloud-based 3D PA iterative reconstruction algorithms.

Moreover, the hierarchical optimization strategy dramatically increases reconstruction efficiency. Results from in vivo rat liver, in vivo rat kidney, and simulated hand vessel reconstructions show that SlingBAG Pro significantly accelerates reconstruction, with the speed advantage becoming more pronounced as the number of detectors increases. For example, with 2006 sensors in the simulated hand vessel experiment, SlingBAG Pro reduces reconstruction time from 8.32 hours to 3.79 hours while maintaining the same reconstruction quality, which demonstrates its remarkable efficiency.

In summary, SlingBAG Pro further optimizes the 3D PAI point cloud iterative reconstruction framework to support arbitrary geometric arrays and significantly improves reconstruction efficiency. This provides a robust algorithmic foundation for high-quality 3D reconstruction, facilitating the miniaturization of photoacoustic imaging devices and the development of wearable array systems.

\section{Methods}\label{sec4}

\subsection{Point cloud-based iterative reconstruction algorithm}\label{subsec4_1}

\begin{figure}[htbp]
\centering
\includegraphics[width=1.0\textwidth]{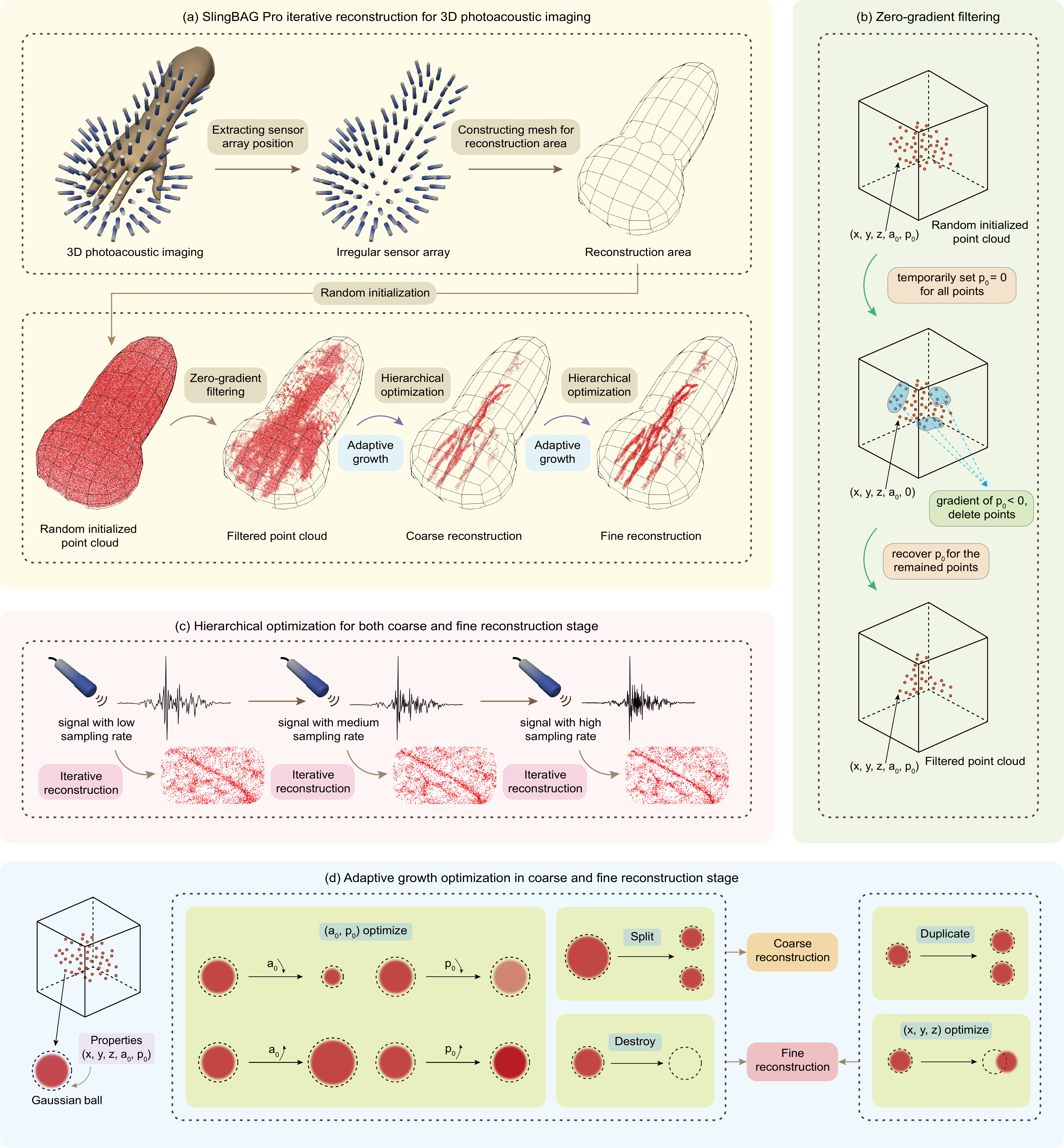}
\caption{The overall framework of SlingBAG Pro iterative reconstruction algorithm for 3D PAI under arbitrary array. (a) The SlingBAG Pro pipeline. (b) Principle of zero-gradient filtering for refinement of point cloud initialization. (c) Hierarchical optimization based on variable sampling rates. (d) Adaptive growth optimization in coarse and fine reconstruction stage.}\label{fig7}
\end{figure}

The Sliding Ball Adaptive Growth (SlingBAG)~\cite{li2024sliding} algorithm provides the first framework for point cloud-based iterative reconstruction for 3D PAI, and the proposed upgraded version, SlingBAG Pro, similarly inherits the point cloud-based iterative approach. In this framework, information of PA sources is stored in the form of a point cloud that undergoes iterative optimization, where the 3D PA scene is modeled as a series of Gaussian-distributed spherical sources. During the iterative reconstruction process, all these Gaussian-distributed sources with specific peak intensity $p_0$ (pressure value), standard deviation $a_0$ (size), and mean value $\mu$ (spatial position $\left(x_s, y_s, z_s\right)$) are used to calculate the predicted PA signals in the position of sensors through the differentiable rapid radiator~\cite{li2024sliding}. By minimizing the discrepancies between the predicted and the actual PA signals, the SlingBAG Pro algorithm iteratively refines the point cloud and ultimately realize the 3D PAI reconstruction by converting the reconstruction result from point cloud into voxel grid (Fig. \ref{fig7}). The point cloud iteration is divided into a coarse reconstruction stage and a fine reconstruction stage for both SlingBAG Pro and SlingBAG algorithms (Fig. \ref{fig7}(d)). In the coarse reconstruction stage, only $p_0$ and $a_0$ are updated, while $\left(x_s, y_s, z_s\right)$ remains fixed; points in the point cloud are adaptively split or destroyed based on thresholds set for pressure and standard deviation. In the fine reconstruction stage, $p_0$, $a_0$, and $\left(x_s, y_s, z_s\right)$ are all updated. In addition to adaptive splitting and destruction, points in the cloud are also adaptively duplicated, increasing the point density to enable more refined reconstruction (Fig. \ref{fig7}(d)).

As an advanced version of the SlingBAG algorithm, SlingBAG Pro provides a reconstruction strategy for arbitrary geometric arrays and greatly enhances the efficiency of point cloud-based iterative reconstruction by incorporating zero-gradient filtering and hierarchical optimization (Fig. \ref{fig7}). Compared to the original SlingBAG, which is limited to cubic reconstruction regions, SlingBAG Pro takes full advantage of point cloud representations: it constructs a polygonal reconstruction mesh according to arbitrary array coordinates provided as input, and uses a ray casting~\cite{haines1994point} method to ensure that all initialized points are located within the irregular mesh, enabling point cloud initialization for any array geometry.

\begin{algorithm}[H]
    \caption{SlingBAG Pro with Zero-Gradient Filtering}
    \label{alg:SlingBAGPro}
    \renewcommand{\algorithmicrequire}{\textbf{Input:}}
    \renewcommand{\algorithmicensure}{\textbf{Output:}}
    \begin{algorithmic}[1]
        \Require $\mathcal{A}$: Arbitrary array geometry (coordinate list)
        \Require $S_0$: Real measured signal
        \Ensure 3D reconstruction result as voxel grid

        \State \textbf{Envelope mesh generation:}
              Construct a closed, polygonal envelope mesh $\mathcal{M}$ based on $\mathcal{A}$
        \State \textbf{Random point cloud initialization:}
              Randomly initialize $N$ points $\{P_i\}$ within $\mathcal{M}$ using ray casting
        \State \textbf{Zero-gradient filtering:}
        \State \quad For each point $P_i$, store random initial pressure $p_i^{\mathrm{init}}$ and temporarily set $p_i = 0$
        \State \quad Perform forward simulation (using $p_i=0$) to compute loss $L_0$
        \State \quad Backpropagate $L_0$ to obtain zero-pressure gradients $g_i = \left. \frac{\partial L}{\partial p_i}\right|_{p_i=0}$
        \State \quad Filter the point set: keep points with $g_i < 0$ to obtain $Z_{0'}$
        \State \quad For retained points, restore $p_i \leftarrow p_i^{\mathrm{init}}$

        \State \textbf{Hierarchical coarse-to-fine optimization:}
        \For{sampling rate $r$ in increasing sequence $\{r_1, r_2, ..., r_n\}$}
            \If{$r$ is $r_1$}    
                \State $Z \gets Z_{0'}$
                \While{not converged}
                    \State Simulate signal $S$ at rate $r$ (using differentiable rapid radiator)
                    \State Compute loss $L = \|S - S_0\|_2^2$
                    \State Backpropagate to update point attributes (e.g., amplitude $a_0$, parameter $p_0$)
                    \State Perform \textit{splitting} and \textit{destroying} operations on $Z$
                \EndWhile
            \Else     
                \While{not converged}
                    \State Simulate signal $S$ at current rate $r$
                    \State Compute loss $L = \|S - S_0\|_2^2$
                    \State Backpropagate to update all coordinates $(x, y, z)$ and point attributes
                    \State Perform \textit{duplicating}, \textit{splitting} and \textit{destroying} operations on $Z$
                \EndWhile
            \EndIf
        \EndFor

        \State \textbf{Point cloud to voxel grid conversion:}
        \For{each point in $Z$}
            \State Convert to voxel grid by point cloud-voxel shader
        \EndFor
    \end{algorithmic}
\end{algorithm}

For the randomly initialized point cloud, SlingBAG Pro applies zero-gradient filtering to drastically reduce the number of points (Fig. \ref{fig7}(b)), so that the filtered point cloud provides a better initial estimate of the PA source distribution and greatly improves reconstruction efficiency. In addition, for both the coarse and fine reconstruction stages, SlingBAG Pro introduces hierarchical optimization based on variable sampling rates (Fig. \ref{fig7}(c)), obtaining a low-resolution point cloud at first by using low-rate supervision, then gradually increasing the sampling rate to yield a high-resolution reconstruction.

\subsection{Zero-gradient filtering for point cloud initialization}\label{subsec4_2}

First, SlingBAG Pro constructs a closed envelope surface based on the spatial coordinates of the input irregular detector array, and defines the reconstruction grid by extending a small distance inward along the normal directions. Using ray casting, the point cloud is then randomly and uniformly initialized within the envelope. For the randomly initialized points in the reconstruction region, we then use zero-gradient filtering to optimize the initial point cloud distribution. Considering that for points located at positions of true photoacoustic sources, there is a tendency for their pressure values to recover from zero to positive values, resulting in negative zero-pressure gradients; in contrast, points with zero-pressure gradients greater than or equal to zero are very unlikely to correspond to actual PA sources. The core process is detailed as below.

In the zero-gradient filtering process, we first save the original initial pressure values \(p_0^{(i)}\) for each source ball \(i\), denoted as:
\begin{equation}
\begin{split}
    p_0^{\mathrm{orig}, (i)} &= p_0^{(i)}.
\end{split}
\end{equation}
To evaluate the contribution of each ball, we temporarily set all initial pressures to zero, i.e.,
\begin{equation}
\begin{split}
    p_0^{(i)} &\leftarrow 0, \qquad \text{for all } i.
\end{split}
\end{equation}
Let $\vec{r}_s$, $\vec{p}_0$ and $\vec{a}_0$ denote the stacked vectors of source locations, initial pressures, and radii, respectively, and let the sensor locations be denoted by $\vec{r}_0$. 
The simulated signal is computed as:
\begin{equation}
\begin{split}
    \mathbf{y}_\mathrm{sim} &= \mathrm{Simulate}(\vec{r}_0,\, \vec{r}_s,\, \vec{p}_0,\, \vec{a}_0).
\end{split}
\end{equation}
The loss function between the simulated and the downsampled real signal is defined by:
\begin{equation}
\begin{split}
    \mathcal{L} &= \left\| \mathbf{y}_\mathrm{sim} - \mathbf{y}_\mathrm{real} \right\|_2^2.
\end{split}
\end{equation}
We then perform backpropagation to compute the gradient of the loss with respect to the initial pressure of each ball:
\begin{equation}
\begin{split}
    g^{(i)} &= \frac{\partial \mathcal{L}}{\partial p_0^{(i)}}.
\end{split}
\end{equation}
Based on the gradient values, we filter the set of balls as follows: if $g^{(i)} \leq 0$, the ball is preserved and its original initial pressure is restored,
\begin{equation}
\begin{split}
    p_0^{(i)} &\leftarrow p_0^{\mathrm{orig}, (i)},
    \qquad \text{if } g^{(i)} \leq 0,
\end{split}
\end{equation}
otherwise, if $g^{(i)} > 0$, the ball is removed from consideration. As a result, after filtering, the set of retained source balls can be written as:
\begin{equation}
\begin{split}
    \left\{ i\,\middle|\, g^{(i)} \leq 0 \right\},
\end{split}
\end{equation}
with the total number of sources updated to match this new set. This procedure ensures that only balls whose pressure increases are expected to decrease the loss are kept for further optimization, while others are effectively pruned from the system, which greatly reduces computational burden and accelerates the iterative process.

\subsection{Hierarchical optimization based on variable sampling
rates and positivity-constrained refinement}\label{subsec4_3}

During the iterative reconstruction process, we introduce a hierarchical downsampling optimization strategy. Specifically, let the full measured supervision signal be denoted as \(\mathbf{y}_\mathrm{real}\), and let the downsampling factor at hierarchical level \(k\) be \(f_k\). At each stage, we downsample the measured signal as
\begin{equation}
\begin{split}
    \mathbf{y}_\mathrm{real}^{(k)} = \mathrm{Downsample}_{f_k}(\mathbf{y}_\mathrm{real}),
\end{split}
\end{equation}
where \(f_k\) decreases stepwise as \(k\) increases, corresponding to progressively higher sampling rates. The simulated signal at each level is
\begin{equation}
\begin{split}
    \mathbf{y}_\mathrm{sim}^{(k)} = \mathrm{Simulate}(\vec{r}_0,\, \vec{r}_s,\, \vec{p}_0,\, \vec{a}_0,\, \Delta t_k,\, N,\, T_k),
\end{split}
\end{equation}
with \(\Delta t_k = f_k \Delta t\) and \(T_k = T / f_k\) reflecting the adjusted temporal resolution for each hierarchical level.

At low sampling rates (\(f_k\) large), the optimization initially produces a coarse, low-resolution result. As the reconstruction proceeds and \(f_k\) is reduced, more high-frequency details are incorporated, enabling the model to progressively recover fine structural information. The loss function at each stage is defined as
\begin{equation}
\begin{split}
    \mathcal{L}^{(k)} = \big\| \mathbf{y}_\mathrm{sim}^{(k)} - \mathbf{y}_\mathrm{real}^{(k)} \big\|_2^2.
\end{split}
\end{equation}
This hierarchical optimization strategy significantly accelerates convergence in the overall reconstruction process.

Both the coarse and fine reconstruction stages in SlingBAG Pro employ this hierarchical optimization approach. In the fine reconstruction stage, the period for changing the downsampling factor \(f_k\) is synchronized with the period of point cloud duplication operations, ensuring that both the spatial sampling density and data fidelity improve together as the optimization progresses.

In the final phase of the fine reconstruction stage, the adaptive density optimization is terminated and a positivity-constrained refinement is performed to enforce physical plausibility. Specifically, we focus on the parameter \(a_0\), which represents the standard-deviation of each Gaussian ball and therefore must be strictly non-negative. To guarantee this constraint while preserving differentiability, we reparameterize \(a_0\) via the softplus function and continue iterative optimization on the unconstrained variable.

Formally, let \(\tilde{a}_0 \in \mathbb{R}\) denote the free optimization variable. The physically valid parameter is obtained through the mapping
\begin{equation}
a_0 = \operatorname{softplus}(\tilde{a}_0) = \log\!\left(1 + e^{\tilde{a}_0}\right).
\end{equation}
The softplus function is a smooth approximation of the rectified linear unit (ReLU) and satisfies
\[
\operatorname{softplus}(x) > 0 \quad \forall x \in \mathbb{R},
\]
thereby ensuring that \(a_0\) remains strictly positive throughout optimization. Compared with hard non-negativity constraints or projection-based methods, this reparameterization avoids gradient discontinuities and enables stable, efficient gradient-based refinement.

During this stage, the loss function remains
\begin{equation}
\mathcal{L}^{\text{final}} = \big\| \mathbf{y}_\mathrm{sim} - \mathbf{y}_\mathrm{real} \big\|_2^2,
\end{equation}
but gradients are backpropagated through the softplus transformation. This final constrained fine-tuning step ensures that the reconstructed parameters not only achieve high data fidelity but also strictly satisfy the underlying physical requirement of a positive standard deviation.

\backmatter

\section*{Acknowledgements}

This research was supported by the following grants: the National Key R\&D Program of China (No. 2023YFC2411700, No. 2017YFE0104200); the Beijing Natural Science Foundation (No. 7232177); the National Natural Science Foundation of China (62441204, 62472213); the Basic Science Research Program through the National Research Foundation of Korea (NRF) funded by the Ministry of Education (2020R1A6A1A03047902).

\section*{Declarations}

\subsection*{Competing interests}

All authors declare no competing interests.

\subsection*{Data availability}

The data supporting the findings of this study are provided within the Article and its Supplementary Information. The raw and analysed datasets generated during the study are available for research purposes from the corresponding authors on reasonable request.

\subsection*{Code availability}

The source code for SlingBAG, along with supplementary materials and demonstration videos, has been made publicly available in the following GitHub repository: \href{https://github.com/JaegerCQ/SlingBAG\_Pro}{https://github.com/JaegerCQ/SlingBAG\_Pro}.

\subsection*{Author contributions}

Shuang Li conceived and designed the study. Yibing Wang and Jian Gao contributed to the design of experiments. Chulhong Kim and Seongwook Choi provided the experimental data. Yu Zhang and Qian Chen contributed to the simulation experiments. Changhui Li and Yao Yao supervised the study. All of the authors contributed to writing the paper.

\section*{Supplementary information}


Please refer to our separate Supplementary Information file for further details.

\bibliography{references}

\end{document}